# Approximation by Quantization

**Vibhav Gogate** and **Pedro Domingos**
Computer Science & Engineering
University of Washington
Seattle, WA 98195, USA
{vgogate,pedrod}@cs.washington.edu

## Abstract

Inference in graphical models consists of repeatedly multiplying and summing out potentials. It is generally intractable because the derived potentials obtained in this way can be exponentially large. Approximate inference techniques such as belief propagation and variational methods combat this by simplifying the derived potentials, typically by dropping variables from them. We propose an alternate method for simplifying potentials: quantizing their values. Quantization causes different states of a potential to have the same value, and therefore introduces context-specific independencies that can be exploited to represent the potential more compactly. We use algebraic decision diagrams (ADDs) to do this efficiently. We apply quantization and ADD reduction to variable elimination and junction tree propagation, yielding a family of bounded approximate inference schemes. Our experimental tests show that our new schemes significantly outperform state-of-the-art approaches on many benchmark instances.

## 1 INTRODUCTION

Many widely used approximate inference algorithms such as mini-bucket elimination (Dechter and Rish, 2003) and the generalized mean-field algorithm (Xing et al., 2003) are essentially *scope-based* approximations. The approximation is invoked when either the factors of the posterior distribution or intermediate functions generated during the execution of a variable elimination algorithm are too large to fit in memory or too time-consuming to compute. Since the time and memory cost of processing a function is exponential in its scope size (in this paper, we consider only discrete graphical models), these schemes reduce complexity by approximating a large-scope function by several small-scope functions. For instance, the generalized mean field algorithm approximates each component $P_i$ of the posterior distribution by a tractable component $Q_i$ defined over a subset of the scope of $P_i$ such that the KL divergence between $Q_i$ and $P_i$ is minimized. The reasons for the popularity of the scope-based approach are obvious; it is a very natural and simple idea, it is easy to implement and its complexity can be easily controlled.

In this paper, we propose a fundamentally different but complementary class of *range-based* approximations: the main idea is to *quantize* a function by mapping a number of distinct values in its range to a single value. When the number of distinct values in the range is reduced, the function becomes more compressible and the time required to manipulate it may decrease substantially. Unfortunately, if we represent functions using tables, namely if we store a real number for every possible configuration of all variables appearing in the function's scope, quantization will be useless because we will not reduce the representation size. In other words, we need structured representations to take advantage of quantization.

Many structured representations have been proposed in literature such as confactors (Poole and Zhang, 2003), sparse representations (Larkin and Dechter, 2003), algebraic decision diagrams (ADDs) (Chavira and Darwiche, 2007), arithmetic circuits (Darwiche, 2003), AND/OR multi-valued decision diagrams (Mateescu et al., 2008) and formula-based representations (Gogate and Domingos, 2010). When a function has a large number of similar values (as a result of quantization or not), the size and compute time of these representations can be exponentially smaller than the tabular representation. Although one can use any of these structured representations or combinations to compactly represent a quantized function, in this paper we propose to use ADDs (Bahar et al., 1993). ADDs are canonical representations of functions, and have many efficient manipulation algorithms. In particular, all inference operations: multiplication, maximization, and elimination can be efficiently implemented using standard ADD operations. Another advantage of ADDs is that there is a large literature on them. This has led to the wide availability of many efficient open source software implementations (e.g., CUDD Somenzi (1998)), which can be leveraged to efficiently and quickly implement the ideas presented in this paper.

Quantization is a general principle that can be applied to a variety of probabilistic inference algorithms. In this paper, we apply it to two standard algorithms: bucket (or variable) elimination (Dechter, 1999) and the junction tree algorithm (Lauritzen and Spiegelhalter, 1988), yielding approximate, anytime and coarse-to-fine versions of these schemes. Just like mini-bucket elimination (Dechter and Rish, 2003) and related iterative algorithms such as expectation propagation (Minka, 2001) and generalized belief propagation (Yedidia et al., 2004), one can view our new schemes as running exact inference on a simplified version of the graphical model. All approximate schemes proposed to date define a simplified model as a low treewidth model.[1] However, treewidth is an overly strong condition for determining feasibility of exact inference (Chavira and Darwiche, 2008). For example, algorithms such as ADD-VE (Chavira and Darwiche, 2007) and formula decomposition and conditioning (Gogate and Domingos, 2010) can solve problems having large treewidth by taking advantage of context-specific independence (or identical potential values) (Boutilier et al., 1996) and determinism. Quantization artificially introduces context-specific independence and thus enables us to define a new class of approximations that take advantage of the efficiency and power of the aforementioned schemes by simplifying the graphical model in a much finer manner.

We present experimental results on four classes of benchmark problems: Ising models, logistics planning instances, networks for medical diagnosis and coding networks. Our experiments show that schemes that utilize quantization and ADD reduction significantly outperform state-of-the-art bounding and approximate inference approaches when the graphical model has a large number of similar probability values or local structure such as determinism and context-specific independence. When the network does not have these properties, our algorithms are slightly inferior to the best-performing state-of-the-art scheme but superior to other state-of-the-art approaches.

The rest of the paper is organized as follows. Section 2 describes background. Section 3 presents quantization. Section 4 presents approximate inference schemes based on quantization. Experimental results are presented in Section 5 and we conclude in Section 6.

## 2 PRELIMINARIES

### 2.1 MARKOV NETWORKS

For simplicity, we focus on Markov networks defined over bi-valued variables. Our approach can be easily applied to multi-valued variables, and other graphical models such as Bayesian networks and Markov logic (Domingos and Lowd, 2009). Let $\mathbf{X} = \{X_1, \ldots, X_n\}$ be a set of bi-valued (Boolean) variables taking values from the domain $\{0, 1\}$ (or {False,True}). A Markov network denoted by $\mathcal{M}$, is a pair $(\mathbf{X}, \mathbf{F})$ where $\mathbf{X}$ is a set of variables and $\mathbf{F} = \{F_1, \ldots, F_m\}$ is a collection of potentials or real-valued Boolean functions of the form $\{0,1\}^k \to \mathbb{R}^+$. Each potential $F_i$ is defined over a subset of variables, denoted by $V(F_i) \subset \mathbf{X}$, also called its scope. The set of values in the range of $F_i$ is denoted by $R(F_i)$. A Markov network represents the following probability distribution:

$$\Pr(\mathbf{x}) = \frac{1}{Z} \prod_{i=1}^{m} F_i(\mathbf{x}_{V(F_i)}) \qquad (1)$$

where $\mathbf{x}$ is a 0/1 truth assignment to all variables $X \in \mathbf{X}$, $\mathbf{x}_{V(F_i)}$ is the projection of $\mathbf{x}$ on the scope of $F_i$ and $Z = \sum_{\mathbf{x}} \prod_{i=1}^{m} F_i(\mathbf{x}_{V(F_i)})$ is the normalization constant, also called the partition function.

In this paper, we will focus on the approximating the partition function $Z$ and the marginal distribution $P(X_i = x_i)$ at each variable $X_i$. Our approach can be easily extended to other problems such as computing the most probable explanation (MPE).

### 2.2 ALGEBRAIC DECISION DIAGRAMS

An algebraic decision diagram (ADD) is an efficient graph representation of a real-valued Boolean function. It is a directed acyclic graph (DAG) in which each leaf node is labeled by a real value and each non-leaf decision node is labeled by a variable. Each decision node has two outgoing arcs corresponding to the true and false assignments of the corresponding variable. ADDs enforce a strict variable ordering from the root to the leaf node and impose the following three constraints on the DAG: (i) no two arcs emanating from a decision node can point to the same node, (ii) if two decision nodes have the same variable label, then they cannot have (both) the same true child node and the same false child node and (iii) no two leaf nodes are labeled by the same real value. ADDs that do not satisfy these constraints are referred to as unreduced ADDs (while those that do are called reduced ADDs). An unreduced ADD can be reduced by merging isomorphic subgraphs and eliminating any nodes whose two children are isomorphic (for details, see Bahar et al. (1993)). ADDs are canonical representations of real-valued Boolean functions, namely, two functions will have the same ADD (under the same variable ordering) iff they are the same.

Figure 1 shows a real-valued Boolean function and its corresponding ADD.

All inference operations (including sum, product, elimination, etc.) can be efficiently implemented using ADDs; their complexity is polynomial in the size of the corresponding ADDs. Unfortunately, the time and memory constants involved in using ADDs are much larger than those involved in using tables. Because of this, ADD-based elimination (and multiplication) may be more expensive, both time-wise and memory-wise, even when they perform fewer numeric operations than table-based elimination (and multiplication). However, when a function has a substantial amount of context-specific independence, the ADD operations can be significantly faster.

---

[1]The only exception we are aware of is the recent work of (Lowd and Domingos, 2010), who compile an arithmetic circuit (which are structured representations similar to ADDs) from dependent samples generated from the posterior distribution. Our approach is very different, and empirically seems to yield much greater speedups (although to date there is no head-to-head comparison in the same domains because an implementation of the Lowd and Domingos scheme is not available).

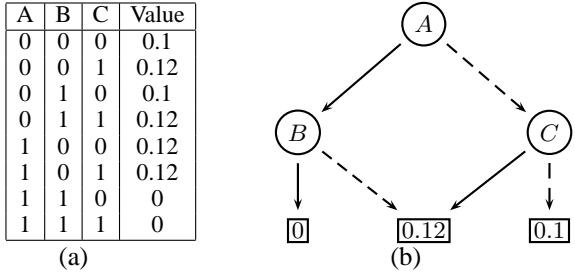

Figure 1: (a) A real-valued Boolean function and (b) its ADD representation. Bold edges in the ADD correspond to true assignments and dashed edges correspond to false assignments. Leaf nodes correspond to the real-values in the range of the function.

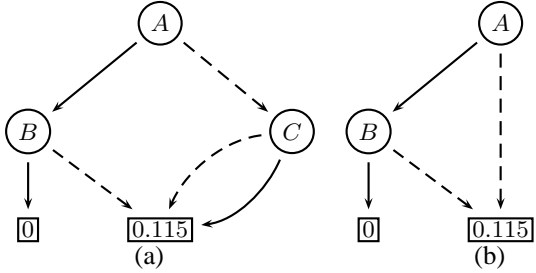

Figure 2: (a) An unreduced ADD obtained by applying quantization $[(0.12, 0.1) \rightarrow 0.115, (0) \rightarrow 0]$ to the ADD given in Figure 1(b) and (b) Reduced ADD obtained from the ADD of (a).

## 3 QUANTIZATION

Quantization is the process of replacing a range of real numbers by a single number. Formally, a quantization function denoted by $\mathcal{Q}$, is a many-to-one mapping from a set **T** to a set **Q** of real numbers, where $|\mathbf{T}| \geq |\mathbf{Q}|$. Let $F$ be a real valued Boolean function, **Q** be a set of real numbers and $\mathcal{Q}$ be a quantization function from $R(F)$ to **Q**. We say that a function $F_Q$ is a quantization of $F$ w.r.t. $\mathcal{Q}$ if $F_Q$ is constructed from $F$ by replacing each value $w$ in the range of $F$ by $\mathcal{Q}(w)$. Quantization may reduce the size of the ADD of a function, but it will never increase it. Formally:

**Proposition 1.** *Let $F_Q$ denote the quantization of $F$ w.r.t. $\mathcal{Q}$. Then, the ADD of $F_Q$ is smaller than or equal to the ADD of $F$.*

Figure 2 demonstrates the effect of quantization on the size of the ADD given in Figure 1(b).

As mentioned in the introduction, the main problem in approximate inference is to find a small bounded function that approximates a large intractable function such that the approximation error is minimized. Assuming that we represent the function using ADDs and approximate using quantizations, we can formalize this problem as follows.

**Quantization Problem:** Given a function $F$, an integer constant $k$ and an error measure $D$ (e.g., KL divergence, mean-squared error, etc.), find a (optimal) quantization $F_Q$ of $F$ such that:

- *Size Constraint:* The size of the ADD of $F_Q$ is less than or equal to $k$.

- *Error Constraint:* There does not exist a quantization $F_{Q'}$ of $F$ such that the size of the ADD of $F_{Q'}$ is less than or equal to $k$ and $D(F, F_{Q'}) < D(F, F_Q)$.

Unfortunately, finding an optimal quantization is extremely hard because the quantization problem is a multi-objective constrained optimization problem. Therefore, we propose the following three heuristics.

Our first heuristic optimizes for error and solves the following relaxation: given an integer $l$ and an ADD $\phi_F$ ($\phi_F$ represents a function $F$) having $t$ leaves, find an ADD $\phi_{F_Q}$ (that represents the quantization $F_Q$ of $F$) having $l$ leaves such that $D(F, F_Q)$ is minimized. This problem can be solved in $O(lt)$ time using dynamic programming and matrix searching (see Wu (1991) for details). Given $l$, the relaxation optimizes $F_Q$ in terms of the error measure $D$ while disregarding the size of the ADD of $F_Q$ (although since $l < t$, $\phi_{F_Q}$ will be smaller than $\phi_F$). To use this heuristic for solving the quantization problem, we have to determine the value of $l$ that will yield an ADD having less than $k + 1$ nodes. To find $l$, we use binary search. We call this heuristic the *min-error* heuristic.

Our second heuristic solves the following relaxation: given an integer $l$ and an ADD $\phi_F$ having $t$ leaves, find an ADD $\phi_{F_Q}$ having $l$ leaves such that there does not exist an ADD $\phi_{F'_Q}$ that has $l$ leaves but fewer nodes than $\phi_{F_Q}$ ($F_{Q'}$ is a quantization of $F$). Unfortunately, this relaxation is much harder to solve than the relaxation that optimizes the error. Therefore, we use the following (heuristic) technique to solve it. As before, we perform a binary search over $l$ starting with $l = t/2$. At each search point, we select a leaf node and merge it with another leaf that shares the largest number of parents with it (ties broken by the relative difference between the leaf values). When two leaves having the same parent are merged, the parent will point to the same leaf node in the new (unreduced) ADD and will be deleted when the ADD is reduced. Notice that the heuristic ignores the error measure $D$ (except when breaking ties) and reduces the ADD size by merging as fewer leaves as possible. Therefore, we call this heuristic the *min-merge* heuristic.

In practice, we can run both heuristics in parallel, compute the error between the original function and the quantized function obtained using each heuristic, and choose the quantized function having the smallest error. We call this heuristic the *min-error-merge* heuristic.

We will evaluate the performance of both the heuristics as well as the combination in the experimental section. Note that when approximations without bounding guarantees are desired, we assign the median value of the merged leaves to the new leaf. When upper (or lower) bounds are desired, we assign the maximum (or the minimum) value instead.

**Algorithm 1**: ABQ($k$)
**Input**: A Markov network $\mathcal{M}$ and a size bound $k$
**Output**: An estimate of the partition function of $\mathcal{M}$
**begin**
    Heuristically select a variable ordering $o = (X_1, \ldots, X_n)$.
    Express each potential of $\mathcal{M}$ as an ADD.
    // Create Buckets
    Let $\mathbf{B}_{X_i}$ be the bucket of $X_i$. Put each ADD in the bucket of its highest ordered variable.
    $Z = 1$
    **for** $i = n$ *downto* 1 **do**
        **repeat**
            // Process the Bucket of $X_i$
            **if** $\mathbf{B}_{X_i}$ *contains only one ADD* $\phi_1$ **then**
                $\phi = \sum_{X_i} \phi_1$
                Put $\phi$ in the bucket of its highest ordered variable. If $\phi$ has no variables then $Z = Z \times \phi$
                Delete $\phi_1$ from $\mathbf{B}_{X_i}$
            **else**
                Heuristically select $\phi_1$ and $\phi_2$ from $\mathbf{B}_{X_i}$.
                $\phi = \phi_1 \times \phi_2$
                Delete $\phi_1$ and $\phi_2$ from $\mathbf{B}_{X_i}$.
                **if** *the size of $\phi$ is greater than* $k$ **then**
                    // Quantization step
                    $\phi_q$ = ADD formed by repeatedly *quantizing and reducing* $\phi$ until its size is less than $k$.
                    Put $\phi_q$ in $\mathbf{B}_{X_i}$.
                **else**
                    Put $\phi$ in $\mathbf{B}_{X_i}$.
        **until** $\mathbf{B}_{X_i}$ *is empty*
    **return** Z
**end**

## 4 APPROXIMATION BY QUANTIZATION

In this section, we apply quantization and ADD reduction to two standard inference algorithms: (i) bucket or variable elimination (Dechter, 1999), and (ii) junction tree propagation (Lauritzen and Spiegelhalter, 1988). Applying quantization and ADD reduction to the former yields a one-pass algorithm for computing the partition function similar to mini-bucket elimination (Dechter and Rish, 2003), and applying it to the latter yields an iterative algorithm that can compute posterior marginal distribution at each variable, similar to expectation propagation (Minka, 2001).

### 4.1 ONE-PASS APPROXIMATION BY QUANTIZATION (ABQ)

Before describing our algorithm, we give background on bucket elimination. Bucket elimination (BE) (Dechter, 1999) is an exact algorithm for computing the partition function. The algorithm maintains a database of valid functions that is partitioned into buckets, one for each variable.

Given an ordering $o$ of variables, the algorithm partitions the potentials of a Markov network by putting each potential in the bucket of the highest ordered variable in its scope. The algorithm operates by eliminating variables one by one, along $o$. A variable $X$ is eliminated by computing a product of all the functions in its bucket, and then summing out $X$ from this product. This creates a new function, whose scope is the union of the scopes of all functions that mention $X$, minus $\{X\}$. The algorithm then deletes the functions involving $X$ (namely the bucket of $X$) from the database of valid functions, adds the newly created function to it and continues. The function (a real number) created by eliminating the last bucket equals the partition function. It is known that the time and space complexity of BE is exponential in the treewidth of the Markov network.

BE assumes tabular representation of functions. It can be easily extended to use ADDs yielding the ADD-BE algorithm, first presented in (Chavira and Darwiche, 2007). In ADD-BE, we represent all functions using ADDs and use ADD operators for elimination and multiplication. Unfortunately, just like BE, it is an exact algorithm and is therefore not scalable to interesting real-world applications.

We propose to make ADD-BE practical by quantizing large ADDs generated during its execution. Algorithm 1 describes the proposed scheme. The algorithm takes as input a Markov network $\mathcal{M}$ and a size bound $k$ and outputs an estimate of the partition function. It is essentially a standard ADD-based bucket elimination algorithm except for the quantization step. Here, given an ADD whose size is greater than $k$, we repeatedly merge its leaf nodes using the heuristics described in the previous section, until its size is smaller than $k$. Note that when $k = \infty$ the algorithm runs full bucket elimination and is equivalent to the ADD-BE algorithm of (Chavira and Darwiche, 2007). Thus, ABQ represents an anytime, anyspace bounded approximation of ADD-BE, controlled by the size bound $k$.

We mention an important technical detail which can positively impact both the complexity and accuracy of ABQ. Notice that after quantizing an ADD, some variables may become irrelevant (for example, variable $C$ is irrelevant to the ADD of Figure 2(b) because it does not appear in any of its internal nodes). Thus, instead of adding the quantized ADD to the current bucket, we can safely transfer it to the bucket of its *highest ordered relevant variable*. Note that variables may also become irrelevant when we multiply two ADDs or eliminate the bucket variable from the ADD. Obviously, we can use the same approach in these cases too and transfer the newly generated ADD to the bucket of its highest ordered relevant variable.

The time and space complexity of Algorithm 1 is summarized in the following theorem:
**Theorem 1.** *The time complexity of $ABQ(k)$ is $O(mk^2)$ where $m$ is the number of potentials and $k$ is the size bound. Its space complexity is $O(\max(mk, k^2))$.*

Algorithm 1 can be easily extended to yield an upper

```
Algorithm 2: IABQ(k)
```
**Input**: A Markov network $\mathcal{M}$ and a ADD size-bound $k$
**Output**: A set of junction tree cliques containing potentials and messages received from neighbors
**begin**
    Construct a junction tree for $\mathcal{M}$
    Let $(e_1, \ldots, e_l)$ be an ordering of edges of the junction-tree for message-passing from leaves to the root
    **repeat**
        **for** $i = 1$ *to* $l$ **do**
            Let $e_i = (u_i, v_i)$
            send-message$(u_i, v_i, k)$
        **for** $i = l$ *downto* $1$ **do**
            Let $e_i = (u_i, v_i)$
            send-message$(v_i, u_i, k)$
    **until** *convergence or timeout*
**end**

**Procedure** `send-message`$(u, v, k)$
**Input**: Cliques $u$ and $v$ of a junction tree and a constant $k$
**Output**: $v$ with the old message (ADD) from $u$ replaced by a new message
**begin**
    Let $(\phi_{u,1}, \ldots, \phi_{u,k})$ be a heuristic ordering of the ADDs currently in the clique $u$ except the message received from $v$
    $\phi_{u,v} = 1$
    **for** $i = 1$ to $k$ **do**
        $\phi_{u,v} = \phi_{u,v} \times \phi_{u,i}$
        **if** *the size of $\phi_{u,v}$ is greater than $k$* **then**
            // Quantization step
            $\phi_{u,v}$ = ADD formed by repeatedly *quantizing and reducing* $\phi_{u,v}$ until its size is smaller than $k$
    Let $sep(u,v) = clique(u) \cap clique(v)$
    $\phi_{u,v} = \sum_{clique(u) \setminus sep(u,v)} \phi_{u,v}$
    Replace the old message from $u$ in $v$ with $\phi_{u,v}$
**end**

(lower) bound on the partition function. All we have to do is ensure that the quantization function $\mathcal{Q}(x)$ used by ABQ is an upper (lower) approximation, namely $\forall w \; F_Q(w) \geq F(w)$ ($\forall w \; F_Q(w) \leq F(w)$). Trivially, a quantization function that replaces each value in the interval by the maximum (minimum) value is an upper (lower) approximation. Formally,

**Theorem 2.** *If all quantizations in Algorithm $ABQ(k)$ use a quantization function $\mathcal{Q}$ satisfying $\forall w \; F_Q(w) \geq F(w)$, then the output of $ABQ(k)$ is an upper bound on the partition function. On the other hand, if $\mathcal{Q}$ satisfies $\forall w \; F_Q(w) \leq F(w)$, then $ABQ(k)$ yields a lower bound on the partition function.*

### 4.2 ITERATIVE APPROXIMATION BY QUANTIZATION (IABQ)

In this section, we will show how to approximate the junction tree algorithm (Lauritzen and Spiegelhalter, 1988) using quantization and ADD reduction. The junction tree algorithm is a message-passing algorithm over a modified graph called the *junction tree*, which is obtained by clustering together variables of a Markov network until the network becomes a tree. The clusters are also called cliques. Each clique is associated with a subset of potentials such that the scope of each potential is covered by the variables in the cliques. The message-passing works as follows. First, we designate an arbitrary cluster as the root and send messages in two passes: from the leaves to the root (inward pass) and then from the root to the leaves (outward pass). The message that a clique $u$ sends to its neighbor $v$ is constructed as follows. In clique $u$, we multiply all the potentials associated with $u$, with all the messages received from its neighbors except $v$, and then eliminate all variables that appear in $u$ but not in $v$. The time and space complexity of the junction tree algorithm is exponential in the maximum cluster size of the junction tree used.

We can construct an approximate version of the junction-tree algorithm using quantization and ADD reduction in a straight-forward manner. Algorithm 2 describes our approach. The algorithm first constructs a junction tree for the Markov network and then sends messages along its edges using the send-message procedure. In the send-message procedure, we send a message from a clique $u$ to clique $v$ by multiplying all ADDs corresponding to the messages (except the one received from $v$) and potentials. Just as in ABQ, if the size of the product ADD is larger than $k$, we recursively apply quantization and ADD reduction until its size is smaller than or equal to $k$. Since, the message propagation is performed on a tree, the algorithm will always converge in two passes (assuming that the quantization heuristics do not change between passes).

IABQ belongs to the class of sum-product expectation propagation (EP) algorithms (see Minka (2001) and Koller and Friedman (2009), Chapter 11) which perform inference by sending *approximate messages*. In practice, we can further improve the accuracy of IABQ by performing *belief-update* propagation instead of sum-product propagation. Belief-update IABQ constructs the message from clique $u$ to clique $v$ by first multiplying, and quantizing if necessary, *all* the incoming messages (*including* the one received from $v$). Then, it projects the resulting factor on $sep(u,v)$ and divides it by the message $\phi_{v,u}$ received from $v$ (thus unlike sum-product IABQ, belief-update IABQ requires the division operation). Belief-update IABQ is not guaranteed to converge in two passes and may not converge at all. However, as we shall see in the experimental section, when it does converge, it often converges very quickly (in 10-30 iterations) and yields highly accurate estimates.

IABQ yields a new class of bounded EP algorithms. Existing bounded EP algorithms use treewidth to determine feasibility of inference. In particular, in the junction tree algorithm, the message between $u$ and $v$ corresponds to a (local) fully-connected (clique) graphical model over the separator $sep(u,v)$. Existing EP algorithms ensure tractability by sending bounded treewidth messages (achieved by introducing new conditional independencies between the sepa-rator variables). IABQ, on the other hand, can create messages having substantially larger treewidth than existing EP algorithms. This is because it uses quantization and ADDs to introduce context-specific independencies between the separator variables.

## 5 EXPERIMENTS

In this section, we compare the performance of ABQ and IABQ with other algorithms from the literature. We also evaluate the impact of various quantization heuristics on accuracy. We experimented with instances from four benchmark domains: (i) logistics planning (Sang et al., 2005), (ii) linear block coding, (iii) Promedas Bayesian networks for medical diagnosis (Wemmenhove et al., 2007) and (iv) Ising models. We implemented our algorithms in C++. We ran our experiments on a Linux machine with a 2.33 GHz Intel Xeon quad-core processor and 16 GB of RAM. We gave each algorithm a memory limit of 2GB and (unless otherwise specified) a time limit of 2 hours. We used the CUDD package (Somenzi, 1998) to implement ADDs. We used the minfill ordering heuristic for constructing the junction tree in IABQ and for eliminating variables in ABQ.

### 5.1 EXPERIMENTS EVALUATING THE BOUNDING POWER OF ABQ

When exact results are not available, evaluating the capability of approximate schemes is problematic because the quality of the approximation (namely how close the approximation is to the exact) cannot be assessed. To allow some comparison on large, hard instances, we evaluate the upper bounding power of ABQ, and compare it with three algorithms from literature: mini-bucket elimination (MBE) (Dechter and Rish, 2003; Rollon and Dechter, 2010), Tree-reweighted Belief Propagation (TRW) (Wainwright et al., 2003) and Box propagation (BoxProp) (Mooij and Kappen, 2008). For a fair comparison, we also compare with our own ADD based implementation of mini-bucket elimination (ADD-MBE). ADD-MBE represents all messages and potentials in the MBE algorithm using ADDs instead of tables (both ADD-MBE and ABQ use the same variable ordering). BoxProp was derived for bounding posterior probabilities and therefore $Z$ is obtained by applying the chain rule to individual bounds on posteriors. We experimented with anytime versions of MBE, ADD-MBE and ABQ. Namely, we start with a crude size-bound, $k = 2$, and increase it progressively by multiplying it by 2, until the algorithm runs out of memory or time. Recall that in ABQ, $k$ bounds the size of the ADD. In MBE, it bounds the size of the new functions created by the algorithm. The results in this subsection were obtained using the min-error-merge heuristic described in Section 3 (we compare the impact of heuristics on accuracy in the next subsection).

Note that almost all the instances that we consider in this subsection are quite hard and the exact value of their partition function is not known (except for the logistics instances). Table 1 shows the results. The first column shows the instance name. The second column shows various statistics for the instance such as the number of variables (n), the domain size (d), the number of potentials (p) and the upper bound on treewidth obtained using the minfill ordering heuristic (w). Columns 3, 4, 5, 6 and 7 show the upper bound on the partition function computed by ABQ, MBE, BoxProp, TRW and ADD-MBE respectively. For each scheme, we also report the relative difference $\Delta$ (defined below) between the log of the best known upper bound $U_{Best}$ and the log of the upper bound $U$ output by the current scheme.

| Instance | (n,d,m,w) | ABQ Z Δ | MBE Z Δ | BoxProp Z Δ | TRW Z Δ | ADD-MBE Z Δ |
|---|---|---|---|---|---|---|
| **Logistics planning** | | | | | | |
| log-1 | (939,2,3785,26) | **5e+20** 0 | 7e+67 2.27 | 9e+108 4.25 | 4e+48 1.34 | 6e+20 0.00129 |
| log-2 | (1337,2,24777,51) | **2e+66** 0 | 2e+242 2.66 | X | X | 1e+68 0.0275 |
| log-3 | (1413,2,29487,56) | **2e+52** 0 | 1e+59 0.128 | X | X | 9e+58 0.127 |
| log-4 | (2303,2,20963,52) | **3e+69** 0 | 4e+356 4.13 | X | X | 1e+90 0.296 |
| log-5 | (2701,2,29534,51) | **3e+110** 0 | 2e+427 2.87 | X | X | 6e+125 0.139 |
| **Medical Diagnosis: Promedas networks** | | | | | | |
| or_chain_100 | (1110,2,1125,59) | **1e-06** 0 | 0.01 0.661 | 8e+03 1.66 | 4e+24 5.17 | 0.009 0.653 |
| or_chain_110 | (1163,2,1176,70) | **2e+05** 0 | 2e+07 0.401 | 4e+21 3.15 | 2e+50 8.65 | 1e+07 0.344 |
| or_chain_120 | (1511,2,1524,76) | **2e+02** 0 | 1e+06 1.52 | 4e+41 16.5 | 7e+53 21.6 | 7e+05 1.45 |
| or_chain_132 | (646,2,717,26) | **4e-09** 0 | 2e-05 0.443 | 0.03 0.819 | 1e+14 2.66 | 4e-07 0.241 |
| **Coding networks** | | | | | | |
| BN_130 | (255,2,511,53) | **6e-52** 0 | 2e-49 0.0499 | 2e-30 0.421 | 3e-45 0.13 | 9e-50 0.0432 |
| BN_131 | (255,2,511,53) | 3e-48 0.0307 | **8e-50** 0 | 4e-33 0.34 | 7e-45 0.1 | 1e-47 0.0427 |
| BN_132 | (255,2,511,53) | **5e-51** 0 | 2e-48 0.0518 | 2e-34 0.33 | 4e-45 0.117 | 6e-49 0.0414 |
| BN_133 | (255,2,511,56) | **1e-46** 0 | 1e-45 0.0181 | 7e-28 0.407 | 1e-43 0.0628 | 5e-45 0.0333 |
| BN_134 | (255,2,511,55) | **1e-48** 0 | 1e-47 0.0174 | 6e-31 0.368 | 3e-44 0.0901 | 6e-45 0.0755 |
| **Ising models** | | | | | | |
| 29x29 | (841,2,1624,29) | 7e+1933 0.00879 | **1e+1917** 0 | 9e+2102 0.097 | 1e+2576 0.344 | 1e+1948 0.0162 |
| 31x31 | (961,2,1860,31) | 3e+2229 0.00115 | **8e+2226** 0 | 5e+2578 0.158 | 9e+2576 0.157 | 1e+2259 0.0144 |
| 33x33 | (1023,2,2112,33) | 1e+2557 0.00527 | **4e+2543** 0 | 4e+2753 0.0826 | 5e+3352 0.318 | 6e+2562 0.00754 |
| 35x35 | (1225,2,2380,35) | 1e+2928 0.0104 | **1e+2898** 0 | 5e+2901 0.00128 | 2e+3761 0.298 | 1e+2933 0.0121 |

Table 1: Table showing the upper bound on the partition function and the log-relative difference Δ for ABQ, MBE and BoxProp, TRW and ADD-MBE. Each algorithm was given a time limit of 2 hours and a memory limit of 2 GB. The best performing scheme is highlighted by bold in each row. 'X' indicates that the algorithm did not return a value.

$$\Delta = \frac{\log(U) - \log(U_{Best})}{\log(U_{Best})} \quad (2)$$

The log-relative difference provides a quantitative measure for assessing the relative approximation quality of the bounding schemes (smaller is better).

**Logistics planning instances** Our first domain is that of logistics planning (the networks are available from (Sang et al., 2005)). Given prior probabilities on actions and facts, the task is to compute the probability of evidence. From Table 1, we can see that ABQ significantly outperforms MBE, ADD-MBE, TRW and BoxProp on all instances (the log-relative difference is quite large). ADD-MBE is much superior to MBE on most instances. This is because the domain has a large amount of determinism and identical probability values which ADD-MBE exploits effectively. ADD-MBE is worse than ABQ suggesting that quantization-based approximations are much better in terms of accuracy than MBE-based approximations.

**Medical Diganosis: Promedas networks** Our second domain is that of noisy-OR medical diagnosis networks generated by the Promedas expert system for internal medicine (Wemmenhove et al., 2007). The global architecture of the diagnostic model in Promedas is similar to the QMR-DT medical diagnosis networks (Shwe et al., 1991). Each network can be specified using a two layer bipartite graph in which the top layer consists of diseases and the bottom layer consists of symptoms. If a disease causes a symptom, there is an edge from the disease to the symptom. The networks are available from UAI 2008 evaluation website (Darwiche et al., 2008). From Table 1, we can see that ABQ is superior to MBE, ADD-MBE, TRW and BoxProp on all instances (notice that for all instances the log-relative difference between ADD based schemes and others is quite large).

**Coding networks** Our third domain is random coding networks from the class of linear block codes (Kask and Dechter, 1999) (the networks are available from the UAI 2008 evaluation website (Darwiche et al., 2008)). From Table 1, we can see that ABQ outperforms MBE, TRW and BoxProp on all instances, except BN_131. On this network, MBE is slightly better than ABQ (because of the overhead of ADDs). On all other networks, ABQ is slightly superior to MBE. ADD-MBE is worse than ABQ on all instances. Again, our results on the coding networks clearly demonstrate that quantization with ADD reduction is a better approximation strategy than MBE.

**Ising models** Our last domain is that of Ising models which are $n \times n$ pair-wise grid networks. They are specified using potentials defined over each edge and each node. Each node potential is given by $(\gamma, 1/\gamma)$ where $\gamma$ is drawn uniformly between $0$ and $1$. The edge potentials are either $(\theta, 1/\theta, 1/\theta, \theta)$ or their mirror image $(1/\theta, \theta, \theta, 1/\theta)$ where $\theta$ is drawn uniformly between $1$ and $\beta$ ($\beta$ is called the coupling strength). We use $\beta = 100$ to generate our networks. From Table 1, we can see that ABQ outperforms BoxProp, TRW and ADD-MBE on these models. However, it is slightly inferior to MBE (notice that the log-relative difference between ABQ and MBE is very small).

Intuitively, ABQ should do well when the graphical model contains many similar or identical probability values in each potential. Ising models are interesting in this respect because they represent the worst possible case for ABQ, with no determinism or context-specific structure at all. Remarkably, ABQ still outperforms BoxProp, TRW and ADD-MBE on these models. In our initial experiments it also outperformed MBE, but it does slightly worse than the latest version, which is the one reported in Table 1. MBE employs sophisticated partitioning heuristics (Rollon and Dechter, 2010) that could also be incorporated into ABQ, and many other optimizations characteristic of a mature system; its good performance relative to ABQ is likely due to these improvements, rather than to the basic algorithm. However, there is in general a tradeoff in using ADDs versus tables, as shown by the ADD-MBE results: ADDs can be exponentially smaller and faster by taking advantage of context-specific independence and determinism, but ADDs incur higher overhead than tables, so the latter may be preferable when there is no structure to exploit.

Overall, we see that that ABQ always outperforms TRW, BoxProp and ADD-MBE, and outperforms MBE on all domains except Ising models. ABQ's advantage increases with the the amount of (approximate or exact) context-specific independence and determinism in the domain, but ABQ still does quite well even when these are absent.

| Instance | min-error-merge | min-error | min-merge |
|---|---|---|---|
| | Z | Z | Z |
| | Δ | Δ | Δ |
| **Logistics planning** | | | |
| log-1 | **5.64e+20** | **5.64e+20** | **5.64e+20** |
| | **0** | **0** | **0** |
| log-2 | **1.52e+66** | 1.42e+69 | 9.08e+66 |
| | **0** | 0.0449 | 0.0117 |
| log-3 | 2.04e+52 | 3.26e+50 | **1.36e+48** |
| | 0.0868 | 0.0495 | **0** |
| log-4 | **2.9e+69** | 4.04e+82 | **2.9e+69** |
| | **0** | 0.189 | **0** |
| log-5 | 2.57e+110 | 3.57e+115 | **8.36e+109** |
| | 0.00444 | 0.0512 | **0** |
| **Medical Diganosis: Promedas networks** | | | |
| or_chain_100 | 1.27e-06 | **6.15e-07** | 3.04e-06 |
| | 0.0508 | **0** | 0.112 |
| or_chain_110 | 1.62e+05 | **1.4e+05** | 6.62e+05 |
| | 0.0125 | **0** | 0.131 |
| or_chain_120 | **242** | 3.99e+04 | 1.38e+07 |
| | **0** | 0.931 | 2 |
| or_chain_132 | **3.72e-09** | **3.72e-09** | **3.72e-09** |
| | **0** | **0** | **0** |
| **Coding networks** | | | |
| BN_130 | **6e-52** | 2e-40 | **6e-52** |
| | **0** | 0.225 | **0** |
| BN_131 | **2.59e-48** | 7.54e-42 | **2.59e-48** |
| | **0** | 0.136 | **0** |
| BN_132 | **4.98e-51** | 6.1e-40 | **4.98e-51** |
| | **0** | 0.22 | **0** |
| BN_133 | **1.48e-46** | 1.74e-41 | **1.48e-46** |
| | 6.55e-06 | 0.111 | **0** |
| BN_134 | **1.46e-48** | 7.26e-41 | 1.15e-47 |
| | **0** | 0.161 | 0.0187 |
| **Ising models** | | | |
| 29x29 | **7.08e+1933** | 2.04e+2002 | **7.08e+1933** |
| | **0** | 0.0354 | **0** |
| 31x31 | **2.95e+2229** | 1.78e+2338 | 2.63e+2264 |
| | **0** | 0.0488 | 0.0156 |
| 33x33 | **1.02e+2557** | 1.91e+2636 | **1.02e+2557** |
| | **0** | 0.031 | **0** |
| 35x35 | **1.15e+2928** | 7.76e+3031 | **1.15e+2928** |
| | **0** | 0.0355 | **0** |

Table 2: Table showing the impact of the three quantization heuristics: (i) min-error-merge, (ii) min-error and (iii) min-merge on the upper bound output by ABQ. For each heuristic, we also report the log-relative error Δ.

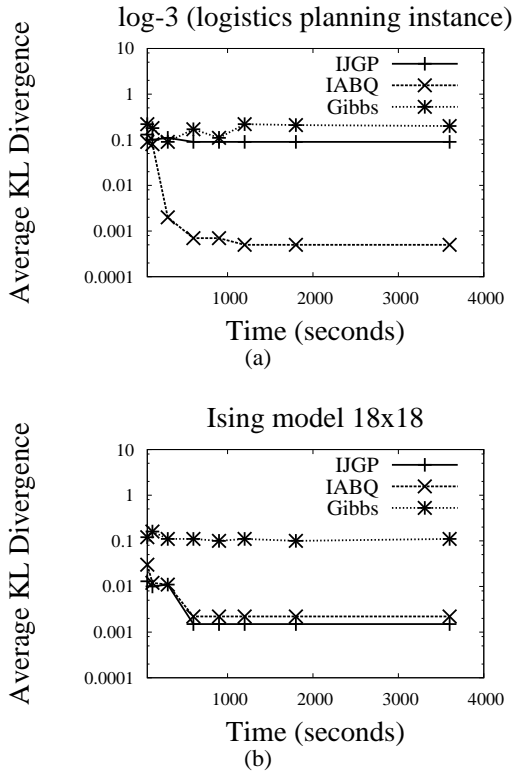

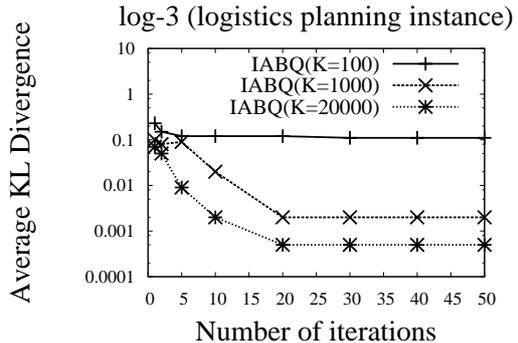

Figure 4: KL divergence vs. Number of iterations for IABQ

Figure 3: KL divergence vs. Time plots for IJGP, Gibbs sampling and IABQ for (a) logistics planning instance log-3 and (b) 18 x 18 Ising model.

## 5.2 EXPERIMENTS EVALUATING THE QUANTIZATION HEURISTICS

In this subsection, we evaluate the performance of the three quantization heuristics described in Section 3. Table 2 shows the results. We can see that the min-error-merge heuristic performs the best overall. The min-merge heuristic is only slightly inferior to the min-error-merge heuristic. The min-error heuristic is inferior to the min-merge heuristic except on the promedas networks. The promedas networks have many similar probability values (approximate context-specific independence) which the min-error heuristic exploits quite effectively. On the other hand, the Ising models represent the worst possible case for the min-error heuristic because the intermediate potentials generated during ABQs execution have almost no similar probability values.

## 5.3 EXPERIMENTS EVALUATING THE ACCURACY OF IABQ

In this subsection, we evaluate the accuracy of belief-update IABQ for computing posterior marginals. We compare IABQ with Iterative Join Graph propagation (IJGP) (Mateescu et al., 2010), a state-of-the-art generalized belief propagation scheme (IJGP won 2 out of the 3 marginal estimation categories at the 2010 UAI approximate evaluation challenge (Elidan and Globerson, 2010)). As a baseline, we compare with Gibbs sampling (Geman and Geman, 1984). We ran both IJGP and IABQ as anytime algorithms. Both algorithms take as input a size parameter which determine their complexity. We vary this parameter starting with its lowest possible value, progressively increasing it until the algorithm runs out of memory or time. We ran each algorithm for 1 hour and gave each algorithm a memory limit of 2GB. Both IJGP and IABQ may or may not converge to a fixed point. Therefore, we ran each for 50 iterations or until convergence, whichever was earlier. Convergence is detected by comparing the absolute difference between messages at the current and previous iteration.

We measure performance using the KL divergence. Let $P(X_i)$ and $Q(X_i)$ denote the exact and approximate marginals of variable $X_i$. Then, the average KL divergence is defined as:

$$KL(P, A) = \frac{1}{|\mathbf{X}|} \sum_{X_i \in \mathbf{X}} \sum_{x_i} P(x_i) \log \left( \frac{P(x_i)}{Q(x_i)} \right)$$

For brevity, we only describe our results for two sample instances: (a) a logistics planning instance, and (b) a 18 x 18 Ising model. Average KL divergence vs. time plots for these instances are given in Figure 3. Our results are consistent with the empirical evidence in the previous subsection. Specifically, when the graphical model has many identical or similar probability values, IABQ dominates IJGP (e.g., on the log-3 instance). However, when the graphical model does not have these properties, IJGP is slightly better than IABQ because of the overhead of ADDs.

Figure 4 shows the impact of increasing the number of iterations on the accuracy of IABQ for different values of the size bound parameter $k$. We can see that IABQ converges to its fixed point in about 10-20 iterations. Its accuracy typically increases with $k$ and with the number of iterations. This shows that the belief-update IABQ performs better than sum-product IABQ (sum-product IABQ is equivalent to running just one iteration of belief-update IABQ).

## 6 CONCLUSION

The most challenging problem in approximate inference is how to approximate a large function that is computation-ally infeasible by a collection of tractable functions. The paper proposes to solve this problem using quantization. Quantization replaces a number of values in the range of a function by a single value, and thus artificially introduces context-specific independence. Conventional tabular representations of functions are inadequate at exploiting this structure. We therefore proposed to use structured representations such as algebraic decision diagrams (ADDs).

We showed how quantization can be applied to two standard algorithms in probabilistic inference, variable elimination and junction tree propagation, yielding two new schemes: (i) A one-pass algorithm that can be used to approximate and bound the partition function and (ii) An iterative algorithm that can be used for approximating posterior marginals. Our new approximate schemes significantly enhance the class of approximations considered by existing algorithms, which constrain their approximations to have low treewidth. By imposing context-specific independencies between variables via quantization, our new algorithms construct structured approximations in the high treewidth space. Our empirical evaluation demonstrates that schemes that employ quantization often yield more accurate results than schemes that do not. Thus approximation by quantization is a promising approach for future investigations.

**Acknowledgements** This research was partly funded by ARO grant W911NF-08-1-0242, AFRL contract FA8750-09-C-0181, DARPA contracts FA8750-05-2-0283, FA8750-07-D-0185, HR0011-06-C-0025, HR0011-07-C-0060 and NBCH-D030010, NSF grants IIS-0534881 and IIS-0803481, and ONR grant N00014-08-1-0670. The views and conclusions contained in this document are those of the authors and should not be interpreted as necessarily representing the official policies, either expressed or implied, of ARO, DARPA, NSF, ONR, or the U.S. Government.